\pdfoutput=1

\documentclass[11pt]{article}

\usepackage{acl}

\usepackage{times}
\usepackage{comment}
\usepackage{latexsym}
\usepackage{enumitem} 
\usepackage[T1]{fontenc}

\usepackage[utf8]{inputenc}

\usepackage{microtype}

\usepackage{microtype}
\usepackage{booktabs}
\usepackage{dsfont}
\usepackage{multirow}
\usepackage{bbding}
\usepackage{caption}
\usepackage{tabularx}
\usepackage{graphicx}
\usepackage{amsmath}
\usepackage{amssymb}
\usepackage{mathtools}
\usepackage{makecell}
\usepackage{amssymb}%
\usepackage{times}
\usepackage{latexsym}
\usepackage{xspace}
\usepackage{csquotes}

\usepackage{xcolor}
\usepackage{color, colortbl}

\definecolor{colorxmark}{RGB}{255, 87, 51}
\definecolor{colorcmark}{RGB}{66, 154, 137}
\definecolor{headcolor}{HTML}{018161}
\definecolor{relationcolor}{HTML}{d95f02}
\definecolor{tailcolor}{HTML}{6560a3}

\newcommand{\Skd}{Symbolic knowledge distillation\xspace}

\newcommand{\skd}{symbolic knowledge distillation\xspace}

\newcommand{\automic}{\textsc{Atomic}$^{\textbf{10x}}$\xspace}
\newcommand{\autotomicbrev}{\textsc{A}$^{\textbf{10x}}$\xspace}
\newcommand{\autotomic}{\textsc{Atomic}$^{\textbf{10x}}$\xspace}
\newcommand{\model}{\textsc{Comet}$^{\textsc{dis}}_{\textsc{til}}$\xspace}

\newcommand{\corpus}{corpus\xspace}
\newcommand{\Corpus}{Corpus\xspace}

\newcommand{\atomicorig}{\textsc{Atomic}\xspace}
\newcommand{\atomic}{$\textsc{Atomic}_{20}^{20}$\xspace}
\newcommand{\atomicbrev}{$\textsc{A}_{20}^{20}$\xspace}
\newcommand{\comet}{$\textsc{Comet}_{20}^{20}$\xspace}

\title{Symbolic Knowledge Distillation:\\from General Language Models to Commonsense Models}

\author{Peter West\textsuperscript{$\dagger$$\ddagger$*} \hspace{.1cm} \textbf{Chandra Bhagavatula}\textsuperscript{$\ddagger$} \hspace{.1cm} \textbf{Jack Hessel}\textsuperscript{$\ddagger$} \hspace{.1cm} Jena D. Hwang\textsuperscript{$\ddagger$} \\
\textbf{Liwei Jiang\textsuperscript{$\dagger$$\ddagger$}} \hspace{.1cm} \textbf{Ronan Le Bras\textsuperscript{$\ddagger$}} \hspace{.1cm} \textbf{Ximing Lu\textsuperscript{$\dagger$$\ddagger$}}  \hspace{.1cm} \textbf{Sean Welleck\textsuperscript{$\dagger$$\ddagger$}} \hspace{.1cm} \textbf{Yejin Choi \textsuperscript{$\dagger$$\ddagger$*}}\\
  \textsuperscript{$\dagger$}Paul G. Allen School of Computer Science \& Engineering, University of Washington\\
  \textsuperscript{$\ddagger$}Allen Institute for Artificial Intelligence\\}

\begin{document}
\maketitle

\begin{abstract}

The common practice for training commonsense models has gone 
\textit{from--human--to--\corpus--to--machine}: 
humans author commonsense knowledge graphs in order to train commonsense models. In this work, we investigate an alternative,  \textit{from--machine--to--\corpus--to--machine}: general language models  
author 
these commonsense knowledge graphs to train commonsense models. 

Our study leads to a new framework, 
\textbf{Symbolic Knowledge Distillation}.
As with prior art in Knowledge Distillation  \cite{Hinton2015KnowledgeDistill}, our approach uses larger models to teach smaller models. A key difference is that we distill knowledge symbolically--as text--in addition to the resulting neural model. We distill only one aspect--the commonsense of a general language model teacher, allowing the student to be a different type of model, a commonsense model. 
Altogether, we show that careful prompt engineering and a separately trained critic model allow us to selectively distill high-quality causal commonsense from GPT-3, a general language model. 

Empirical results demonstrate that, for the first time, a \emph{human-authored} commonsense knowledge graph is surpassed by our \emph{automatically distilled} variant in all three criteria: quantity, quality, and diversity. In addition, it results in a neural commonsense model that surpasses the teacher model's commonsense capabilities despite its 100x smaller size. We apply this to the \atomicorig resource, and will share our new symbolic knowledge graph and commonsense models\footnote{We will share this following the anonymity period. We have permission from OpenAI to release GPT-3 generations}.

\end{abstract}

\section{Introduction}
\label{sec:intro}
\begin{figure}
    \centering
    \includegraphics[width=0.98\linewidth]{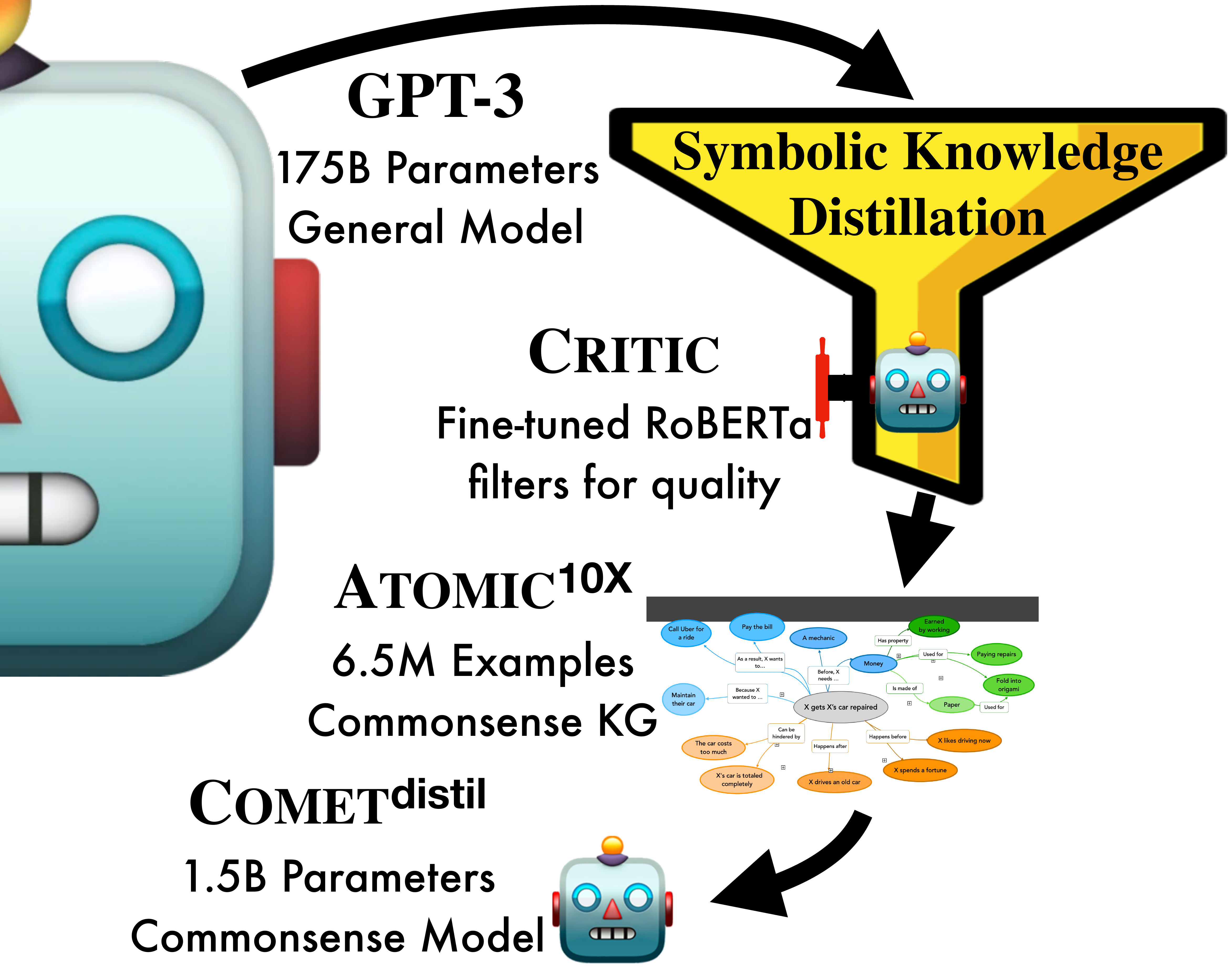}
    \caption{\textbf{\Skd}  extracts the commonsense from the large, general language model GPT-3, into 2 forms: a large commonsense knowledge graph 
    \textbf{\autotomic}, and a compact commonsense model \textbf{\model}. The quality of this knowledge can be controlled and improved by adding a \textbf{critic} model, making GPT-3 a stronger teacher.
    }
    \label{fig:1}
\end{figure}

Prior works have suggested that pre-trained language models possess limited understanding of commonsense knowledge \cite{Merrill2021ProvableLO, csqa20, davis2017causal} 
despite otherwise stellar performance on leaderboards. As a result, symbolic commonsense knowledge graphs \cite{speer2017conceptnet,sap2019atomic,hwang2020comet} and corresponding neural representations \cite{Bosselut2019COMETCT,hwang2020comet,Zhang2020ASERAL} have supplemented past models with commonsense capabilities. This has enabled diverse downstream applications, including interactive learning through a conversational interface \cite{Arabshahi2021ConversationalMR}, 
persona- and affect-aware conversation models \cite{Kearns2020AWI}, figurative language understanding \cite{chakrabarty-etal-2020-r,chakrabarty2021mermaid}, story telling \cite{ammanabrolu2020automated} and fantasy games \cite{Ammanabrolu2021HowTM}.





The common practice for commonsense knowledge graph construction sees humans spell out as many pieces of knowledge as possible. This pipeline goes \textit{from--human--to--\corpus--to--machine}, with commonsense models trained from human-authored knowledge graphs. Yet, high-quality, human-authored knowledge is expensive to scale, limiting coverage; this motivates an alternative: \textit{from--machine--to--\corpus--to--machine}. Prior efforts toward automatic commonsense knowledge graphs have resulted in considerably lower quality than human-written data \cite{hwang2020comet,Zhang2020ASERAL}, which in turn leads to less reliable neural models \cite{hwang2020comet}. Broad literature consistently shows machine-authored knowledge graphs underperform human-authored graphs \cite{etzioni2011open, Mitchell2015NeverEndingL, Bollacker2008FreebaseAC}. 


In this work, we propose \textbf{\Skd}, a new conceptual framework towards high-quality automatic knowledge graphs for commonsense, leveraging state-of-the-art models and novel methodology. Most prior art for automatic knowledge graph construction extracts knowledge from raw text \cite{Bhakthavatsalam2020GenericsKBAK, Zhang2020TransOMCSFL, Zhou2020TemporalCS, Zhang2020ASERAL, causalbank}. 
In contrast, our approach is motivated by knowledge distillation \cite{Hinton2015KnowledgeDistill} wherein a larger teacher model transfers knowledge to a compact student model (\S\ref{subsec:knowledgedistillation}). 
Our method differs from prior knowledge distillation in key ways: we distill a symbolic knowledge graph (i.e., generated text) in addition to a neural model, and we distill only a selective aspect of the teacher model. This selectively allows the student model to be of a different type (commonsense model), compared to the teacher (general language model), enriching the scope of distillation. An added benefit is that knowledge distilled as text is human readable: it can be understood and evaluated. 

A general language model--GPT-3 in our case--is an imperfect commonsense teacher on its own, and the ability to evaluate distilled knowledge is useful in improving it. We empirically demonstrate that, by training a separate critic model to judge symbolic generation quality, a more precise teacher can be defined. Knowledge from this critical teacher is higher quality--even exceeding human-authored knowledge. Yet even before training a critic, our study makes the unexpected finding that the student model surpasses the commonsense of GPT-3, our knowledge source.

To test \skd against the \textit{human--to--\corpus--to--machine} paradigm, we compare with \atomic \cite{hwang2020comet}, which is a human-authored commonsense knowledge graph. We find that \textbf{\autotomic}, our machine-generated \corpus, exceeds the human generated \corpus in \emph{scale, accuracy,} and \emph{diversity} with respect to 7 commonsense inference types that we focus on in this study. The resulting commonsense model, \textbf{\model}, not only surpasses the human-trained equivalent \comet, but is also smaller, more efficient, and produces commonsense at a higher accuracy than its own teacher--GPT-3. 

\Skd offers a promising new role for general language models, as commonsense knowledge sources, and humans, as small-scale evaluators to train critic models rather than authors of commonsense knowledge. Our work demonstrates that humans and LMs can be effective collaborators for curating commonsense knowledge graphs and training efficient and performant commonsense models. 

\section{Overview and Key Findings}
\label{sec:overview}

\begin{figure*}
    \centering
    \includegraphics[width=1.0\linewidth]{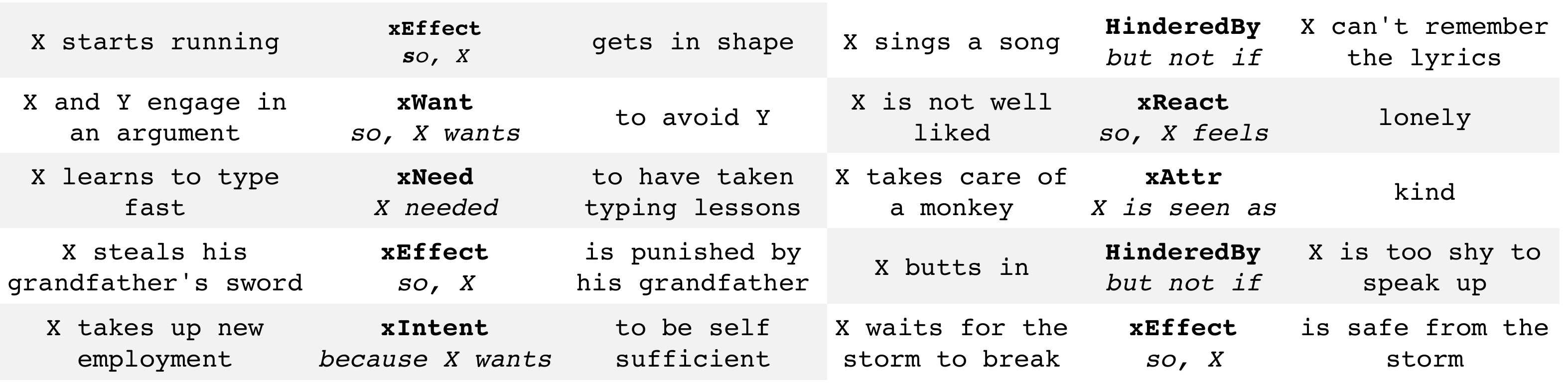}
    \caption{Example \textbf{automatically generated} \atomicorig triples from our \autotomic commonsense knowledge graph. Each example includes a generated \textbf{event}, \textbf{relation} (with natural language interpretation), and generated \textbf{inference}.
    }
    \label{fig:examples}
\end{figure*}

Throughout our work, we describe the \textit{machine--to--corpus--to--machine} methodology of \skd. We first go \textit{machine--to--corpus} (\S\ref{sec:verbalization}), by decoding from GPT-3, then improve our knowledge with a specialized critic model (\S\ref{sec:purification}), and finally distill this knowledge into an efficient commonsense model (\S\ref{sec:distillation}), going \textit{corpus--to--machine}. Throughout this process, we evaluate against a human knowledge source, comparing our automatic knowledge graph \autotomic and commonsense model \model to the human-authored \atomic and resulting model \comet \cite{hwang2020comet}.


\subsection{Symbolic Knowledge Distillation}
\label{subsec:knowledgedistillation}

Our proposed methodology parallels knowledge distillation  \cite{Hinton2015KnowledgeDistill}, a method for
compressing a large or complicated teacher distribution $P_t$ into a smaller/simpler student distribution $P_s$. Key to knowledge distillation\footnote{In its simplest case, with temperature set to 1.0} is the notion of minimizing the cross-entropy between $P_t$ and $P_s$:

\begin{equation}
    H(P_t,P_s) = - \sum_{y \in Y} P_t(y) \log P_s(y)
    \label{eq:kd}
\end{equation}

\noindent
Knowledge is transferred to the student by encouraging it to match teacher predictions. \citet{Hinton2015KnowledgeDistill} apply this to conditional classification: for each training input, $P_t$ and $P_s$ are model predictions over label set $Y$. Typically $Y$ is a tractable set, over which this sum can reasonably be calculated.

For distilling the knowledge of generative models, we can think of an unconditional language model (LM e.g. GPT-3) as $P_t$. This makes $Y$ the set of all strings, over which LMs define probability. Unfortunately $Y$ is an exponential set, intractable to sum over in Eq~\ref{eq:kd}. \citet{Kim2016SequenceLevelKD} address this problem by simply taking the mode of $P_t$ over $Y$, truncating most of the teacher distribution to the most likely sequence and discarding information.

Instead, we consider a sampling-based interpretation of the same objective:

\begin{equation}
    H(P_t,P_s) = \mathop{\mathbb{E}}_{ y \sim P_t(y) } [-\log P_{s}(y)]
    \label{eq:SKD}
\end{equation}

\noindent
which exactly equals the cross-entropy of Eq~\ref{eq:kd}, at the limit under pure sampling from $P_t$.\footnote{A useful consequence of this framing is that access to the full model distribution is not required. Our experiments (\S\ref{sec:verbalization}) use GPT-3, for which the distribution is \textbf{not available}, thus our method is applicable while knowledge distillation is not.}

Yet distilling \emph{all knowledge} from the teacher may not be desirable--our work is specifically focused on distlling  commonsense knowledge from GPT-3. The ideal teacher $P_t$ is a commonsense expert, but GPT-3 can approximate such a teacher, off-the-shelf, via prompting. This ability to select information is one explicit benefit of the sampling-based interpretation of Eq~\ref{eq:SKD}: while Eq~\ref{eq:kd} uses continuous logits over existing data, sampling gives discrete control over transferred information, by selecting which samples are elicited and used. For the general language model GPT-3, We encourage domain/quality with prompting, and sample truncation \citep{holtzman2019curious}. We call this the \emph{loose teacher} $P_t^L$--knowledge is generated and transferred from GPT-3, but without critical assessment of correctness (\S\ref{sec:verbalization}). 

In fact, sampling knowledge in Eq~\ref{eq:SKD} offers even more control, as generations can be individually interpreted and judged. Given an indicator function $A(x)$ for which knowledge $x$ is \emph{correct}, we can define a stronger teacher model. Using a Product of Experts \cite{hinton2002training}  between the loose teacher $P_t^L$ and and the critic $A(x)$, we define a \emph{critical teacher}: 

\begin{equation}
    P_t(x) \propto P_t^L(x|p)\cdot A(x)
\end{equation}
\noindent
In practice, $A(x)$ is a textual classifier learned on human judgements, 1 for knowledge predicted to be correct and 0 otherwise. Thus, the critic gives control over the correctness and confidence of the knowledge that is transferred (\S\ref{sec:purification}). 

\subsection{Key Findings}

Applying \skd in practice results in promising and surprising findings:

\paragraph{1. Learning symbolic knowledge from language models can be framed as a symbolic extension to knowledge distillation.} In \S\ref{subsec:knowledgedistillation}, we describe learning commonsense as a symbolic extension to knowledge distillation, with GPT-3 a knowledge source. We elaborate on this process with positive results in \S\ref{sec:verbalization},\ref{sec:purification}, and \ref{sec:distillation}.

\paragraph{2. \Skd constructs a high quality knowledge graph at scale.} Our method naturally yields a machine-generated commonsense knowledge graph, which can achieve impressive quality (\S\ref{sec:purification}), beyond that of human-authored data. An effective critic which filters incorrect generated knowledge is key.

\paragraph{3. A critical teacher results in a higher quality student.} In \S\ref{sec:purification}, we show that making the teacher more critical results in higher quality knowledge, even as it reduces the scale of knowledge transferred. This demonstrates that \emph{quality} matters, not just \emph{quantity}, as higher quality knowledge results in a higher quality commonsense model in \S\ref{sec:distillation} despite smaller scale data.

\paragraph{4. Critical teacher or not, a student can outperform the knowledge source.} In \S\ref{sec:distillation}, we show the unexpected result that all student models exceed the quality of GPT-3, the knowledge source. 

\paragraph{5. Machines can win over humans for automatic knowledge graph construction.} In \S\ref{sec:purification} and \S\ref{sec:distillation}, we show that machine generated knowledge and the resulting commonsense model can outperform their equivalents that use a human knowledge source. Our symbolic knowledge exceeds humans at scale, quality, and diversity. The resulting commonsense model achieves the most accurate commonsense KG completions.

\section{Machine-to-\Corpus Verbalization}
\label{sec:verbalization}

\label{subsec:verbalization}

\Skd begins by going \textit{machine--to--\corpus}, i.e. generating many commonsense facts, which results in a commonsense knowledge graph. \S\ref{subsec:knowledgedistillation} frames this as sampling to estimate the knowledge distillation objective--a student commonsense model learns from the generations of a teacher (GPT-3).

We start with a \emph{loose teacher}, transferring knowledge by prompted generation with truncated sampling alone--this is in contrast to the \emph{critical teacher} (\S\ref{sec:purification}) which explicitly judges and filters the generated samples. The loose teacher uses few-shot prompting as in \citet{gpt3}. We use a few-shot template:

\begin{center}
\resizebox{.7\linewidth}{!}{
\begin{tabular}{l}
  \textbf{\texttt{ <TASK-PROMPT> }} \\
  \textbf{\texttt{ \textcolor{headcolor}{<EX$_1$-INP>}\textcolor{tailcolor}{<EX$_1$-OUT>} }}
  \\
  \ldots \\
    \textbf{\texttt{ \textcolor{headcolor}{<EX$_{N-1}$-INP>}\textcolor{tailcolor}{<EX$_{N-1}$-OUT>} }}\\
    \textbf{\texttt{ \textcolor{headcolor}{<EX$_N$-INP>} }}
\end{tabular}
} \end{center}

\noindent where \texttt{<EX$_{i}$-INP>}/\texttt{<EX$_{i}$-OUT>} are human-authored, natural language \atomicorig entries, and \texttt{<TASK-PROMPT>} is a description of the problem.
Given such a prompt, GPT-3 generates the \emph{missing piece}, output \texttt{<EX$_{N}$-OUT>} for input \texttt{<EX$_{N}$-INP>}, following the pattern of earlier examples (1 to N-1). We find important aspects for producing high-quality commonsense knowledge:

\begin{itemize}[leftmargin=*,topsep=0pt,itemsep=-1ex,partopsep=1ex,parsep=1ex]
    \item Examples should be numbered. e.g. \texttt{<EX$_{5}$-INP>} might begin with "5)" to indicate it is the 5th example.
    \item The format of \texttt{<EX$_{i}$-INP>} and \texttt{<EX$_{i}$-OUT>} should linguistically imply the relationship between them. See below for examples.
    \item \texttt{<TASK-PROMPT>} can be used to give extra specification to complicated problems.
\end{itemize}

\subsection{Data: \atomicorig}
\label{subsec:atomicdata}

We demonstrate \skd on the \atomicorig \textit{if-then} resource \cite{sap2019atomic}.
This follows an event-relation-inference (triple) 
format. The \corpus links \textit{events} (e.g. \textit{X attacks Y}) to relations, e.g. \textbf{HinderedBy} which describes what might hinder an event. For a relation/event, the goal is to generate a resulting inference, e.g. \textit{X attacks Y} \textbf{HinderedBy} \textit{X is restrained}. 

Of the 23 relations from the most recent version--\atomic--we limit our investigation to 7 relations that correspond to \emph{causal} commonsense knowledge: 
\textbf{xAttr} (how X is perceived after \textit{event}), \textbf{xReact} (how X reacts in response to \textit{event}), \textbf{xEffect} (what X does after \textit{event}), \textbf{xIntent} (X's intent in \textit{event}), \textbf{xWant} (what X wants after \textit{event}), \textbf{xNeed} (what X needed for \textit{event} to take place) and \textbf{HinderedBy}. 
We describe how \textbf{verbalization} is applied to \atomicorig data in 2 steps: generating underlying events (heads), then full examples (inference given event).

\subsection{Event Generation}
\label{subsubsec:eventgen}

Events are context-free premises in \atomicorig involving \texttt{PersonX} (and sometimes a second \texttt{PersonY}) in various scenarios. These events form heads in knowledge graph triples. We generate events by filling in the elements of our template: 



\begin{center}
\resizebox{.98\linewidth}{!}{
\begin{tabular}{l}
  \textbf{\texttt{\textcolor{headcolor}{1.} \textcolor{tailcolor}{ Event: X overcomes evil with good}}} \\
  \textbf{\texttt{\textcolor{headcolor}{2.} \textcolor{tailcolor}{Event: X does not learn from Y}}} \\
  \ldots \\
  \textbf{\texttt{\textcolor{headcolor}{10.} \textcolor{tailcolor}{Event: X looks at flowers}}} \\
  \textbf{\texttt{\textcolor{headcolor}{11.}}} \\
\end{tabular}
} \end{center}

\noindent
The format is simple, as events are generated \textit{unconditionally}.
We use 100 high-quality events from the \atomic corpus for our prompt, selected to avoid grammatical or logical errors, and minimize semantic overlap. We randomly sample 10 of these seed events for each generation batch, resulting in randomized prompts. We use nucleus sampling ($p=0.9$) \cite{holtzman2019curious}, and presence/frequency penalties of 0.5 from the GPT-3 interface. We generate 165K unique events using the 175B-parameter Davinci model\footnote{the largest available version of GPT-3} from \citet{gpt3} (human-authored \atomic contains only 6.2K events).

\subsection{Inference Generation}
\label{subsec:inf-gen}

Generating \atomicorig inferences requires reasoning about events and relations together.
We design verbalization templates fo reach relation, with iterative design and small-scale verification by the authors\footnote{See Appendix~\ref{app:prompts} for full prompts.} e.g. we prompt the \textbf{xNeed} relation as follows:
\begin{displayquote}

\textbf{\texttt{\small What needs to be true for this event to take place?}}

\ldots

\textbf{\textcolor{headcolor}{\texttt{\small Event <i>: X goes jogging}}}
\textbf{\texttt{\small \textcolor{headcolor}{Prerequisites: For this to happen,} \textcolor{tailcolor}{\textbf{X needed to wear running shoes}}}}

 \ldots

\textbf{\textcolor{headcolor}{\texttt{\small Event <N>: X looks at flowers}}}
\textbf{\texttt{\small \textcolor{headcolor}{Prerequisites: For this to happen,}}}
\end{displayquote}

\noindent
The language of this template implies the relation-specific task, both "Prerequisites:" and beginning with "for this to happen" suggest the \textbf{xNeed} relation. As well, we include an xNeed-specific \texttt{<TASK-PROMPT>}. 
We use 10 few-shot examples for each prompt.\footnote{We also replace anonymous names (``X'') with sampled generic names as this improved quality, See Appendix~\ref{app:prompts}. Once generation is complete, we substitute in generic markers (``X'') for the final dataset.}

For each event/relation (165K X 7) we generate 10 inferences with the 
Curie GPT-3 model\footnote{for the largest, Davinci, 12M generations is computationally/monetarily intractable.} 
and earlier hyperparameters. Removing duplicate and degenerate (e.g. fewer than 3 characters) 
generations yields 6.46M \atomicorig-style data triples (examples in Figure~\ref{fig:examples}). We call this \autotomic, as it contains an order of magnitude more triples than \atomic for the 7 relations we study. 

\subsection{Evaluating a Generated Commonsense Knowledge Graph}
\label{subsec:verbalization-eval}

Machine generation enables a large scale of unique generations at a much lower cost than human-authored knowledge (Table~\ref{tbl:relations-size}), but what kind of examples are produced by GPT-3, and how does it differ from knowledge produced by humans? In this section, we conduct an in-depth analysis to answer these questions.

\paragraph{Lexical Differences: Diversity and Uniqueness}

Recent work finds that machine generations can be repetitive and lack diversity \cite{Welleck2020NeuralTG,holtzman2019curious}; one way generated knowledge may differ from human-authored is less creative word choice, diversity, or more repetition.

To test this, we begin with lexical diversity (i.e. unique words used, Table~\ref{tbl:tail-tokens-by-relation}). While there is variation by relation, the diveristy of \autotomic actually exceeds \atomic here, 5.2M unique words to 1.5M.
In addition, it contains significantly more strictly unique generated inferences (Table~\ref{tbl:tail-tokens-by-relation}, unique tails).

\paragraph{BLEU Soft Uniqueness.} Exact match (above) fails to capture the notion of \textit{similar} text. Following the intuition of self-BLEU \cite{zhu2018texygen}, we define \textit{soft uniqueness} to describe diversity of generations in a corpus. An inference $x$ is softly-unique if: 

$$BLEU_2(C,x) < 0.5$$

\noindent
where $C$ is the set of inferences for a given input (in our case, event + relation), and 0.5 is an empirical threshold. To find soft-uniqueness of a corpus,
we iteratively remove examples until all are softly unique, i.e. low mutual lexical overlap; higher diversity means more such examples (thus a larger softly unique corpus is preferable). 
Softly-unique corpus sizes are given in Table~\ref{tbl:ATOMIC-human} (``Size (div)''). \autotomic has a smaller \emph{fraction} of softly-unique examples than \atomic, yet it contains many more such examples. \autotomic contains 4.38M such examples (full size 6.5M) vs. \atomic, which has 560K (full size 600K).


\begin{table}[!t]
\setlength{\tabcolsep}{4pt}
\begin{center}
\small
\begin{tabular}{lrr}
\toprule
\textbf{Relation}   & \atomic & \automic \\
\midrule
\texttt{HinderedBy} & 77,616 & \textbf{1,028,092} \\
\texttt{xNeed} & 100,995 & \textbf{760,232} \\
\texttt{xWant} & 109,098 & \textbf{730,223} \\
\texttt{xIntent} & 54,839 & \textbf{965,921} \\
\texttt{xReact} & 62,424 & \textbf{1,033,123} \\
\texttt{xAttr} & 113,096 & \textbf{884,318} \\
\texttt{xEffect} & 90,868 & \textbf{1,054,391} \\
\midrule
\textbf{Total Count} & 608,936 & \textbf{6,456,300} \\
\textbf{Est Total Cost} & \textasciitilde\$40,000 & \textbf{\textasciitilde\$6,000} \\
\textbf{Est Cost Per Triple} & \textasciitilde\$0.06 & \textbf{\textasciitilde\$0.001} \\
\bottomrule
\end{tabular}
\end{center}
\caption{Number of unique triples with the given relation, $|(\cdot,\texttt{relation},\cdot)|$. 
The estimated cost for \automic{} comes at a fraction of a conservative estimation for \atomic crowdsourcing costs.
}
\label{tbl:relations-size}
\end{table}

\begin{table}[!t]
\setlength{\tabcolsep}{4pt}
\begin{center}
\footnotesize
\begin{tabular}{lrrrrrr}
\toprule
& \multicolumn{2}{c}{} & \multicolumn{2}{c}{\textbf{Unique}} & \multicolumn{2}{c}{\textbf{Unique}}
\\
& \multicolumn{2}{c}{\textbf{Length}} & \multicolumn{2}{c}{\textbf{ Tokens (K)}} & \multicolumn{2}{c}{\textbf{ Tails (K)}} \\
\cmidrule(lr){2-3} \cmidrule(lr){4-5} \cmidrule(lr){6-7}
& \textbf{\atomicbrev} & \textbf{\autotomicbrev} & \textbf{\atomicbrev} & \textbf{\autotomicbrev} & \textbf{\atomicbrev} & \textbf{\autotomicbrev}\\
\texttt{xWant} & 4.69 & 5.16 & 322 & 784 & 69 & 152\\
\texttt{xAttr} & 1.42 & 2.73 & 15 & 21 & 11 & 8\\
\texttt{xEffect} & 3.92 & 4.66 & 216 & 864 & 55 & 185 \\
\texttt{xIntent} & 4.59 & 5.92 & 136 & 800 & 30 & 135 \\
\texttt{xNeed} & 4.51 & 5.97 & 289 & 1378 & 64 & 231 \\
\texttt{xReact} & 4.03 & 1.77 & 48 & 5 & 12 & 2 \\
\texttt{HinderedBy} & 7.93 & 7.49 & 522 & 1775 & 290 & 874 \\ 
\midrule
Events & 5.20 & \textbf{5.32} & 109 & \textbf{881} & 6.2 & \textbf{165}\\
\bottomrule
\end{tabular}
\end{center}
\caption{
Average length, total unique tokens and total unique examples (in K, i.e. 1000s) by relation type and in events (bottom row) from \atomic (\atomicbrev) and \automic (\autotomicbrev).}
\label{tbl:tail-tokens-by-relation}
\end{table}





\paragraph{Model-based Diversity Measurement.} 
Lexical notions of diversity reward differences in surface form, which may not always reflect diversity of \textit{information}, only format. Thus, we next study information-theoretic measures for diversity. Intuitively, diverse information should be less predictable, or higher entropy. With GPT-2 XL models finetuned on \atomic and \autotomic (\S\ref{sec:distillation}) we estimate \textbf{entropy}--roughly, how difficult it is for a model to capture the corpus information (Table~\ref{tbl:entropy}). This is 4 times higher for \autotomic, suggesting more content from a modeling perspective. We also estimate \textbf{cross-entropy}--how well a model trained on one corpus describes the other. From \autotomic to \atomic, this is 9.31, only 2 points higher than its entropy suggesting \atomic is describable with information from \autotomic. 
In reverse, this is 41.48 suggesting much of \autotomic is not captured by \atomic--\autotomic is surprising given only information from \atomic.







\begin{table}[!t]
\setlength{\tabcolsep}{4pt}
\begin{center}
\footnotesize
\scalebox{.88}{
\begin{tabular}{rr|rr|rr}
\toprule
 \multicolumn{2}{c|}{\textbf{Entropy}}  & \multicolumn{2}{c|}{\textbf{Cross Entropy}}
 & \multicolumn{2}{c}{\textbf{KL Divergence}} \\
\midrule
 \multicolumn{2}{c|}{$H(D_{1})=1.27$}&  \multicolumn{2}{c|}{$H(D_{1}, D_{2})=9.31$}
 & \multicolumn{2}{c}{$D_{KL}(D1||D2)=8.04$} \\
\midrule
 \multicolumn{2}{c|}{$H(D_{2})=7.80$}&  \multicolumn{2}{c|}{$H(D_{2}, D_{1})=41.48$}
 & \multicolumn{2}{c}{$D_{KL}(D_2||D_1)=33.68$} \\
\bottomrule
\end{tabular}}
\end{center}
\caption{
Entropy, cross-entropy, and divergence of \atomic\ ($D_1$) and \automic\ ($D_2$).
}
\label{tbl:entropy}
\end{table}


\paragraph{Human Evaluation of Quality.}
\label{subsec:atomichuman}

Perhaps most importantly, we study the \textit{quality} of knowledge in each \corpus. We conduct human evaluation with Amazon Mechanical Turk. 3 annotators rate each triple resulting in ``accepted'', ``rejected'' or ``no judgement''. We evaluate 3000 examples\footnote{this ensures at least 1000 after filtering by the critic \S\ref{sec:purification})} from \autotomic, and 1000 from \atomic (Table~\ref{tbl:ATOMIC-human}). We find Fleiss' kappa \cite{fleiss1971measuring} of 40.8 indicating moderate agreement \cite{landis1977measurement}, and 90.5\% accuracy agreement. We require workers meet an Amazon Mechanical Turk qualification for annotation quality based on past commonsense evaluations. We compensate workers $\$0.17$ per task, which we estimate require 30 seconds. Further details and task template are in appendix \S\ref{app:human_eval}.

For the \emph{loose teacher}, consider the top row of \autotomic in Table~\ref{tbl:ATOMIC-human} (other rows add the critic \S\ref{sec:purification}). \autotomic exceeds \atomic in scale, but is somewhat less acceptable by human raters--by roughly 8 percentage points. Yet, the larger scale of \autotomic implies a significantly higher \emph{number} of accurate examples. Increasing the proportion of these is the main objective of the critic (\S\ref{sec:purification}).

\paragraph{How do Knowledge Sources Compare?} 
To understand the robustness of our approach, we assess other language models as the knowledge source (i.e. loose teacher): GPT-J \cite{gpt-j} and T5-11B adapted for language modelling \cite{Lester2021ThePO}. We substitute both for GPT-3 as in \S\ref{subsubsec:eventgen},\ref{subsec:inf-gen}, generating a small-scale corpus to evaluate. We conduct human evaluation on 1000 examples as above (Table~\ref{tbl:ATOMIC-human}). Both models attain roughly 72\% accuracy, 6 points below GPT-3 (78.5). This suggests strong potential, but higher quality from GPT-3. We explore this further in Appendix~\ref{app:alternate-teachers}.

\begin{table}[!t]
\setlength{\tabcolsep}{4pt}
\begin{center}
\footnotesize
\begin{tabular}{l|ccc|c|c}
\toprule
\textbf{Corpus}   & Accept & Reject & N/A & Size & Size (div) \\
\midrule
\atomic             & 86.8 &    11.3 &  1.9 & 0.6M & 0.56M \\
\midrule
\autotomic{} &  78.5 &    18.7 &  2.8 & \textbf{6.5M} & \textbf{4.38M} \\
 &    88.4 &     9.5 &  2.1 & 5.1M & 3.68M\\
 (critic$_\text{low}$) &    91.5 &     6.8 &  1.7 & 4.4M & 3.25M \\
 &    95.3 &     3.8 &  1.0 & 3.0M & 2.33M \\
(critic$_\text{high}$) &    \textbf{96.4} &     2.7 &  0.8 & 2.5M & 2.00M \\
\midrule
+ GPT-J & 72.0 & 27.6 & 0.4 & - & -\\ 
+ T5-11B LM & 71.7 & 26.9 & 1.4 & - & -\\
\bottomrule
\end{tabular}
\end{center}
\caption{Attributes of \automic and \autotomic{} (row 2) including the critic model (\S\ref{sec:purification}, rows 3 - 6) with various filtering cutoffs. Accept and Reject are by majority human vote unless any mark N/A. Size is in unique examples\footnotemark. The highest precision corpus is \autotomic{} with (critic$_\text{high}$), but multiple versions surpass \atomic. We also include alternate models (GPT-J and T5-11B) as the loose teacher.
}
\label{tbl:ATOMIC-human}
\end{table}

\footnotetext{Size of \atomic is given as the number of comparable datapoints, i.e. those with the same relations as \autotomic{}.}

\section{Making the Teacher More Critical}
\label{sec:purification}

Symbolic knowledge distillation requires a strong teacher model to maximize the quality of the generated knowledge graph and resulting student model (\S\ref{sec:distillation}).
While the \emph{loose teacher} (GPT-3 alone) results in a viable commonsense knowledge graph, evaluation shows this isn't a perfect commonsense teacher.
Thus, we multiply in a \emph{critic model}, to filter lower-quality knowledge, \emph{correcting the teacher} (\S\ref{subsec:knowledgedistillation}). With modest supervision (a small-scale human evaluation) we train a classifier to predict and discriminate unacceptable examples. We multiply this with the \emph{loose teacher} \S\ref{sec:verbalization}, creating a \emph{critical teacher} product of experts. In practice this means filtering \autotomic to create new corpora that are higher quality, yet still larger scale than human-authored \atomic. 



\label{subsec:purification}

\paragraph{Training a knowledge critic}

We gather a training set of \textit{correct vs. incorrect} human judgments on a randomly-sampled set of 10K entries of \autotomic, as in \S\ref{subsec:atomichuman} but with one annotation per example. We take a (random) train/dev/test split of 8k/1k/1k. While this step requires human annotation, humans take on the role of high-level supervisors here--critiquing a small number of generations rather than authoring the entire knowledge graph as in previous work. Indeed, the cost/complexity of this step is similar to a typical human evaluation, making it far cheaper/easier than eliciting human-authored knowledge in past work. 

We train binary classifiers (critics) for human acceptability using RoBERTa-Large \cite{liu2019roberta}. We find pretraining on MNLI results in the best model in terms of precision and recall, and we suggest this technique for future studies. We give more detail in Appendix~\ref{app:critic}, including baselines. Our best model vastly improves the accuracy of \autotomic (Table~\ref{tbl:ATOMIC-human}), demonstrating that a small amount of human supervision can consistently help to correct GPT-3's mistakes. 

\paragraph{Size-accuracy trade-off} Using our critic to filter knowledge results in a natural trade-off between size and accuracy. We test several cutoffs for \autotomic, i.e. confidence at which the critic rejects examples. We report human-measured accuracy (Accept/Reject column Table~\ref{tbl:ATOMIC-human}) following \S\ref{subsec:atomichuman}. We compare the loose teacher (unfiltered) to critical teachers.
Discarding 20\% of instances that the critic judges as least acceptable (reducing corpus size from 6.5M to 5.1M),
\autotomic's accuracy rises 78.5 $\rightarrow$ 88.4; human-authored \atomic contains 600K entries at 86.8\% accuracy. Reducing to total size to 2.5M examples (38\% of full size), we attain 96.4\% accuracy, nearly 10 points above \atomic while still 4X larger.




\label{subsec:purificationeval}

\paragraph{What gets filtered out?}

We qualitatively identify two types of filtered triples: 1) \emph{logical misalignments}, events/inferences joined in an inconsistent manner. Recognizing these requires understanding events-inference interactions, e.g., \textit{X cannot find his shirt \textbf{as a result X} is wearing a shirt}; 2) \emph{awkward phrasings}, in which events/inferences are individually incoherent e.g. \textit{PersonX has a fire in the bath}--resulting triples are invalid as the event is implausible.

\begin{table}[t]
\begin{center}
\begin{tabular}{l | ccccc}
\toprule
& Random & Inf & Event & EMAP & Full \\
\midrule
AP & 79.3 & 81.9 & 86.2 & 87.1 & \textbf{94.0}\\
\bottomrule
\end{tabular}
\end{center}
\caption{Average Precision for ablated critic models. The critic not only filters  \emph{awkward phrasings} which can be identified by either the event (\textbf{Event}) or inference (\textbf{Inf}) in isolation (EMAP only identifies these), but also \emph{logical misalignments}, which require modeling interactions between event/inference, i.e. the full critic (\textbf{Full}). 
}
\label{tbl:purification_ablation}
\end{table}



To understand what is filtered, 
we ablate the critic (Table~\ref{tbl:purification_ablation}): our full model is compared to a random predictor, event-only model, and inference-only model. We also compare to an EMAP \cite{hessel2020emap} version, i.e. an ensemble of event and inference-only, without interactions between event/inference (needed for \emph{logical misalignments}).

We find GPT-3 produces both independent awkwardly-phrased events/inferences (filtered by X-only models) and logical misalignments. The classifier, trained on validated knowledge triples, helps in both cases. The EMAP of our full model (identifies only awkward phrasings) achieves 87\% AP, and our full model (which additionally identifies logical misalignments) improves to 94\% AP.


\paragraph{Does filtering hurt diversity?}
One concern is that the critic may keep only similar ``safe'' examples, lacking novelty. We repeat our diversity analysis (\S\ref{subsec:verbalization-eval}) for critical corpora (Table~\ref{tbl:ATOMIC-human}, ``Size (div)'', higher=better). As we filter, we surprisingly observe proportionally \emph{more} diverse examples:
full \autotomic has a diverse subset 68\% of its size; rising to 80\% with the most extreme filtering. One possibility is that GPT-3 gravitates towards common sentence structures for inconsistent knowledge. These would be recognizable to the critic, and removing them would increase both quality and diversity. This surprising result warrants further study.


\section{\Corpus-to-Machine: Distillation}
\label{sec:distillation}
The final step of \skd trains a compact model on the generated natural language knowledge graph. Our base model is GPT2-XL trained on all of \automic: we denote this model by \model. We additionally train the model on critical versions of \automic --$\text{crit}_\text{low}$  
denotes training on the corpus achieving 91.5\% accuracy, and $\text{crit}_\text{high}$ on the 96.4\% accuracy corpus. Models are trained for 1 epoch, with default parameters using the Huggingface Transformers library \cite{Wolf2019HuggingFacesTS}. 



\label{subsec:distillation}

\subsection{Evaluating a Symbolically Distilled Model}
\label{subsec:distillationeval}

\begin{table}[!t]
\setlength{\tabcolsep}{4pt}
\begin{center}
\footnotesize
\begin{tabular}{l|c|ccc}
\toprule
CKG Completion& Train Corpus &  &  & \\
Model  & Acc  & Accept & Reject & N/A  \\
\midrule
GPT2-XL zero-shot & - & 45.1 &     50.3 &       4.6 \\
GPT-3 & - & 73.3 &     24.1 &       2.6 \\
\comet & 86.8 & 81.5 &     16.3 &       2.2 \\
\midrule
\model & 78.5  &    78.4 &     19.2 & 2.4 \\
\ $+\text{critic}_\text{low}$ & 91.5 &  82.9 &     14.9 &       2.2 \\
\ $+\text{critic}_\text{high}$ & 96.4  & \textbf{87.5} &     10.2 &       2.3 \\ 
\bottomrule
\end{tabular}
\end{center}
\caption{Model performance on knowledge base completion, measured by human judgement. Inferences are generated on held-out events from \atomic. Models besides GPT-3 use GPT-2 XL architecture. \model with a strong critic ($+\text{critic}_\text{high}$) achieves the highest acceptance rate overall--87.5.
}
\label{tbl:COMET-human}
\end{table}

Evaluation follows past work \cite{hwang2020comet, Bosselut2019COMETCT, sap2019atomic} testing the ability of models to do knowledge base completion, i.e. generating inferences for test events, specifically from the \atomic test set. We use
human evaluation\footnote{We find Fleiss' kappa \cite{fleiss1971measuring} of 47.1 for acceptance, indicating moderate agreement. \cite{landis1977measurement}, and accuracy agreement of 88.7\%.} following Section~\ref{subsec:atomichuman}, on 1000 inputs (event + relation), with results in Table~\ref{tbl:COMET-human}. We compare to the GPT2-XL-based \comet model trained on human-generated \atomic, and GPT-3 using the same generation method as \S\ref{sec:verbalization}--in effect, comparing the student \model to the \emph{loose teacher} GPT-3. We omit the \emph{critical teacher} (GPT-3 + critic), which is not assured to produce an inference for each input, as the critic may reject all tails for some inputs. We also compare to zero-shot GPT2-XL \cite{Radford2019LanguageMA} using the same methodology (Table~\ref{tbl:COMET-human}).

\paragraph{How does \model compare to GPT-3?} In knowledge distillation, the student model often deteriorates in performance \cite{Hinton2015KnowledgeDistill, Kim2016SequenceLevelKD} compared to its teacher. Comparing our base teacher--GPT-3--to the simplest version of \model (top-row \model of Table~\ref{tbl:COMET-human}) surprisingly shows the student \textit{surpasses} GPT-3, the model that generates its training data\footnote{The slight difference in acceptability for GPT-3 from Table~\ref{tbl:ATOMIC-human} is likely due to variance in raters between rounds of evaluation, and a different distribution of events--Table~\ref{tbl:ATOMIC-human} uses generated events while Table~\ref{tbl:COMET-human} uses events from \atomic.}. We posit that the superior performance of \model may have to do with mistakes of GPT-3 being filtered by verbalization and training of GPT-2, and possibly the focus of \model on one commonsense domain while GPT-3 covers a more general domain. We leave further study of this effect for future work. 

\paragraph{How does \model compare to human knowledge?} While \model \emph{without the critic} is slightly outperformed by \comet in terms of accuracy, this reverses with the critic. For both cutoffs tested, \model surpasses \comet, with more filtering resulting in a wider gap. 

\paragraph{Usefulness of \model} For on-demand inference, where a single high quality inference for some input event/relation is required, \model is the \textbf{best available model}: the most performant version surpasses \comet by 5 points and GPT-3 by over 10. The critical teacher (GPT-3 + critic) yields a more accurate \textit{\corpus}, but may filter all inferences for an input, giving no output. 

\paragraph{Limits and Future Work} The success of symbolic knowledge distillation is a first step--demonstrating superior performance to human authoring on the commonsense relations tested here. No aspect of our approach is specific to these relations, yet further work is needed to explore the feasibility of generation for other aspects of commonsense and knowledge, beyond these relations, to concepts like physical or temporal commonsense.

\section{Related Work}

\paragraph{Commonsense Knowledge Graphs (CKG)} CKGs provide knowledge for commonsense reasoning.
%
Some are manually constructed, e.g. \atomicorig \cite{sap2019atomic, hwang2020comet}. ConceptNet \cite{speer2017conceptnet} contains taxonomy and physical commonsense, authored by humans or compiled from such sources. 
Some CKGs are automatically constructed: TransOMCS \cite{Zhang2020TransOMCSFL} extracts 18.48M tuples from syntactic parses
and CausalBank \cite{causalbank} extracts 314M cause-effect pairs by pattern-matching. In contrast, we \textit{generate} commonsense.




\paragraph{Extracting Knowledge from LMs}

Past work uses models for automatic knowledge graph completion \cite{Bosselut2019COMETCT, hwang2020comet, causalbank}. Yet, models are trained on \textit{existing} resources; \autotomic is generated without these. Other works mine factual/commonsense knowledge directly from off-the-shelf LMs \cite{petroni2019language, davison2019commonsense, Xiong2020Pretrained}, but not resulting in the quality at scale of \autotomic. 

\paragraph{Knowledge Distillation}

Other works use knowledge distillation \cite{Hinton2015KnowledgeDistill} for generation.  \cite{Sanh2019DistilBERTAD} follow a label smoothing formulation, while \citet{Kim2016SequenceLevelKD} follow a similar formulation to us (\S\ref{subsec:knowledgedistillation}), but use the mode of the teacher distribution rather than sampling. Our work is unique in distilling \textit{specific} information (commonsense) from a general language model.




\paragraph{Data Generation}

While manual dataset creation is expensive and complex 
\cite{schwartz-etal-2017-effect, Agrawal2018DontJA, tsuchiya-2018-performance,  Bras2020AdversarialFO}, crowdsourcing is the most popular method for goal-oriented, high quality/coverage datasets. 

Past automatic data mainly use extractive approaches, e.g. syntactic parsing \cite{Zhang2020TransOMCSFL} or pattern matching \cite{causalbank} from unstructured text \cite{Lehmann2015DBpediaA, buck2014n}. 
These scale, but are noisy and limited in format--\atomicorig knowledge will not appear simply in natural text. 
Some works explore automatic data synthesis/expansion by finetuning LMs on existing labeled data \cite{anabytavor2019notenoughdata,Papanikolaou2020DAREDA,kumar-etal-2020-data, yang-etal-2020-generative}, 
but are limited by data quality.

\section{Conclusions}

We introduce \skd, a \emph{machine--to--corpus--to--machine} pipeline for commonsense that does not require human-authored knowledge--instead, using machine generation. Knowledge is transferred from a large, general model to a compact commonsense model, through a commonsense corpus--yielding a commonsense knowledge graph and model. Our resulting symbolic knowledge graph has greater scale, diversity, and quality than human authoring. \skd offers an alternative to human-authored knowledge in commonsense research.

\section*{Acknowledgments}
This work was funded in part by the Natural Sciences and Engineering Research Council of Canada (NSERC) (funding reference number 401233309), DARPA MCS program through NIWC Pacific (N66001-19-2-4031), and the Allen Institute for AI.

\section*{Ethical Considerations} One aspect of our work with the potential for ethical pitfalls is large-scale generation from pretrained language models, in constructing \autotomic{}. Recent work \cite{stochasticparrots} has highlighted the risks of models trained on massive text resources, as GPT-3 \cite{gpt3} is, which we use for generation. Indeed, open generations from pretrained language models can often contain harmful, biased, or offensive aspects. We argue here that this risk is largely mitigated in our work, mainly due to the narrow and constrained nature of our generations. The goal of our work is characterising simple and generic anonymous situations, specifically in terms of commonsense causes and effects. We ensure generations are focused on these topics through careful prompting, which we found to be quite effective at keeping these generations on-topic. As such, the potential for harmful generation is very low; indeed, in a manual inspection of 100 generated examples, we found none that were significant harmful, besides one that contained adult content. 

A related concern is the potential for large models and training sets to make automated oppression or exploitation possible, for instance in surveillance or generating fake news. As above, we argue that the generic, commonsense nature of our data and models makes this concern less relevant here. Our data does not contain any information directly related to these harmful domains (e.g. social media or fake news generation). While our data may assist machines in understanding basic situations, this is unlikely to be useful for harmful models given the simplicity of our data and still-flawed commonsense capabilities of even the most advanced models.

Finally, we note that we ensure fair and generous compensation for all human evaluators we hire through Amazon Mechanical Turk. Based on our estimates of time required per task, we ensure that the effective pay rate is at least \$15 per hour.



\bibliography{anthology,custom}
\bibliographystyle{acl_natbib}

\clearpage

\appendix



\label{sec:appendix}
\section{Human Evaluation Details}
\label{app:human_eval}

\begin{figure*}
    \centering
    \includegraphics[width=0.9\linewidth]{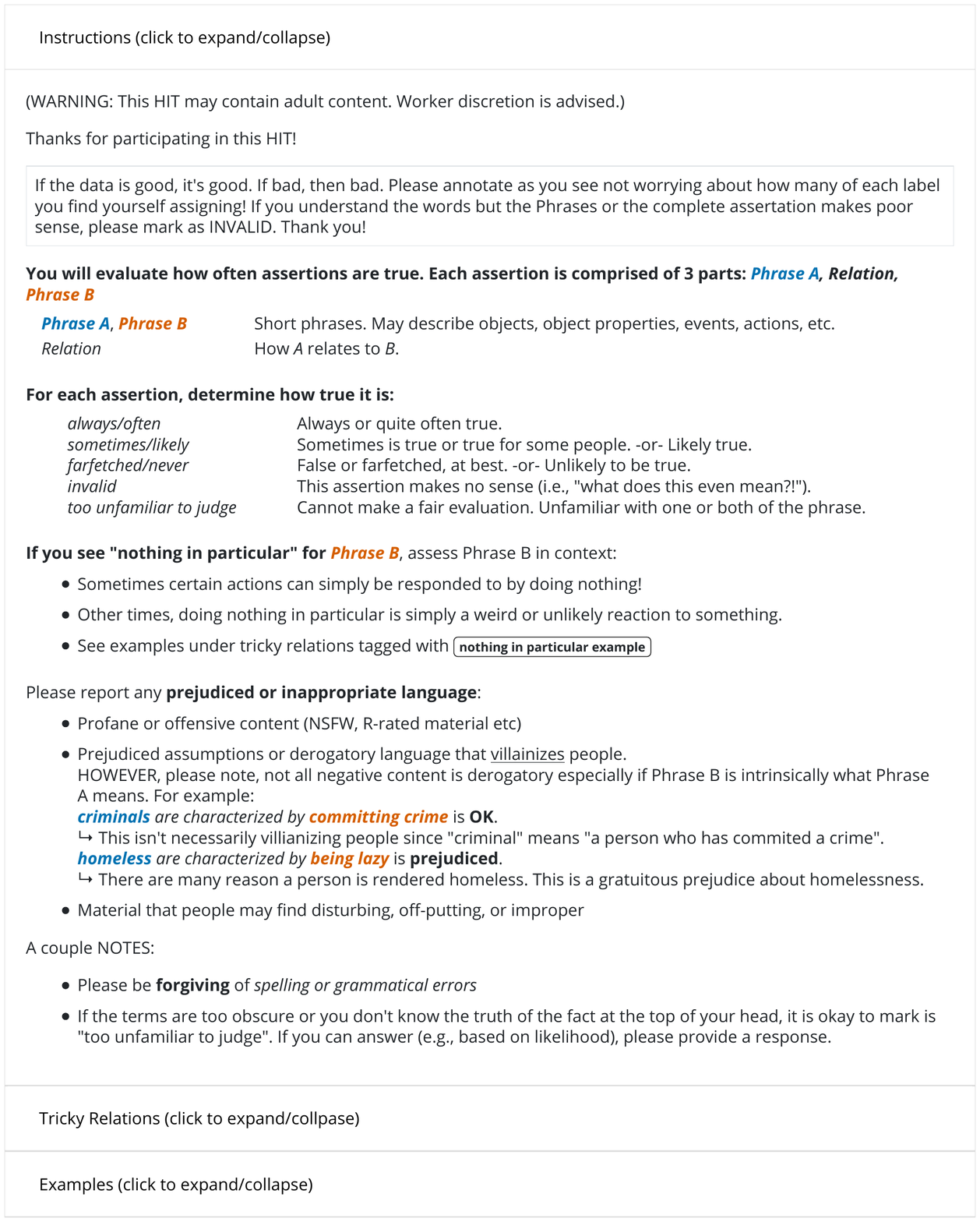}
    \caption{Page 1 of template used for human evaluation.
    }
    \label{fig:template_0}
\end{figure*}

\begin{figure*}
    \centering
    \includegraphics[width=0.9\linewidth]{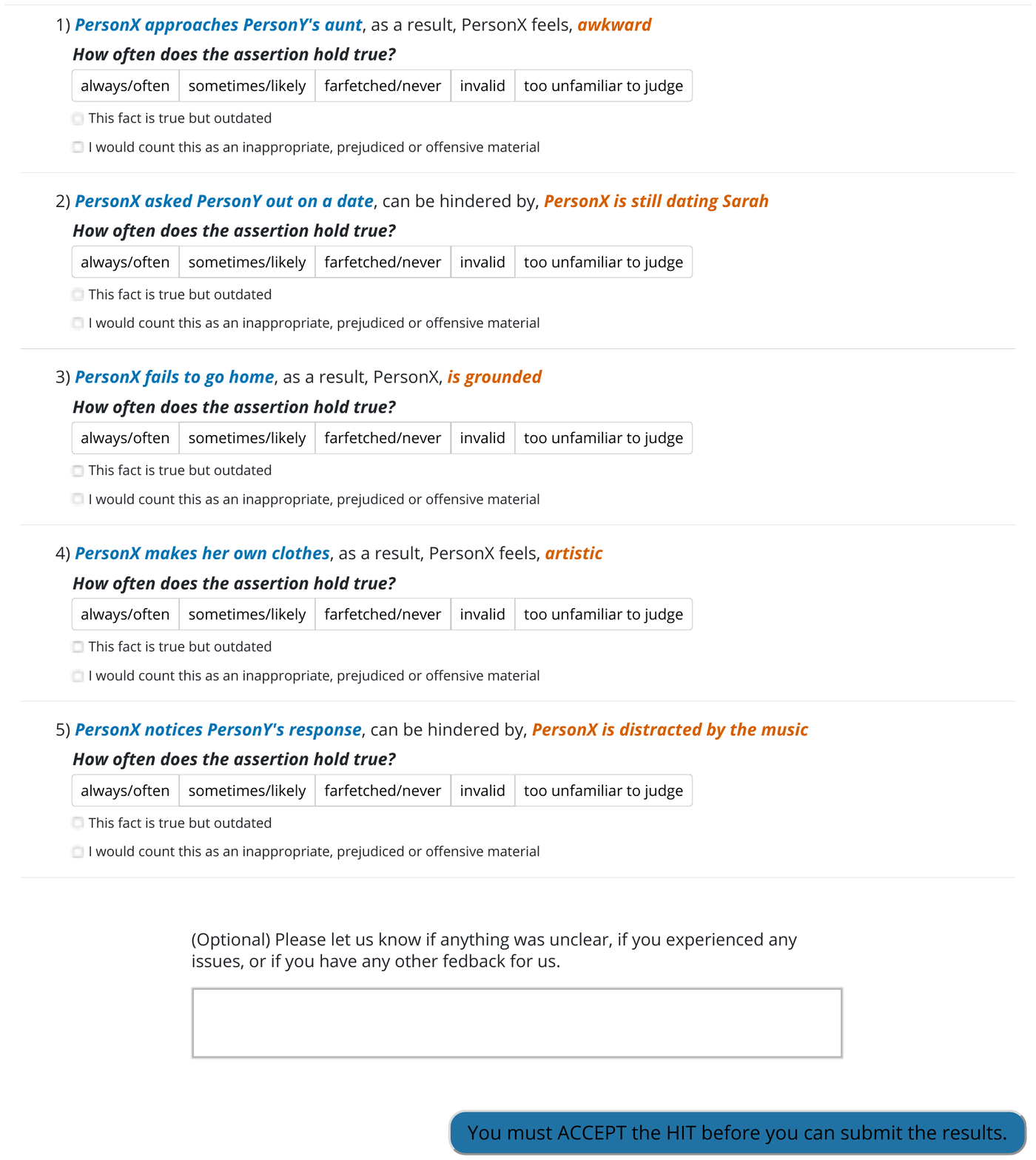}
    \caption{Page 2 of template used for human evaluation.
    }
    \label{fig:template_1}
\end{figure*}

We conduct human evaluations on Amazon Mechanical Turk using the template of Figures~\ref{fig:template_0},\ref{fig:template_1}. Workers are presented with \atomicorig{}-style triples, replacing relations with natural language templates (e.g. \textit{HinderedBy} becomes ``can be hindered by''). 3 annotators rate each triple, with options for acceptability: ``always/often'', ``sometimes/likely'', ``farfetched/never'', ``invalid'', or ``too unfamiliar to judge''. The first two are considered ``accepted'', the second two ``rejected'' and the final is ``no judgement''. For reporting acceptance rates, and training a critic model, we only distinguish between ``accepted'' and not ``accepted''.

Workers are compensated \$0.17 per task (i.e. completing all questions in the evaluation template Figures~\ref{fig:template_0},\ref{fig:template_1}). We estimate an upper bound of 30s to complete a single task, which gives an hourly rate of \$20.4. Workers are selected based on an Amazon Mechanical Turk qualification, specifically filtering for workers with high accuracy on past knowledge base triple evaluations. We follow the same setup for all evaluations, besides number of annotators. This setup is shown to result in consistent and reliable annotations, with an inter-annotator agreement given by Fleiss' kappa \cite{fleiss1971measuring} of 40.8 when evaluating with 3 annotators, in \S\ref{subsec:verbalization-eval}.

\section{Using Alternate Models as Knowledge Sources}
\label{app:alternate-teachers}

One natural question that arises from the strong performance of symbolic knowledge distillation is whether other sources of knowledge (i.e. language models) would similarly benefit from this method. In this section, we particularly measure the capacity of other language models to serve as the ``loose teacher'' which generated the base knowledge of the resulting corpus.

We expand our study beyond GPT-3 here (the model used in our work), to include 2 contemporary large language models, GPT-J \cite{gpt-j} and T5-11B \cite{Lester2021ThePO} finetuned for language modelling. For knowledge generation (verbalization) we follow the same procedure as \S\ref{sec:verbalization} along with simple adjustments to improve quality. We are investigating the effect of the critic on knowledge precision here, so we also include \atomic to probe the usefulness of automatic filtering for human-authored knowledge.

For each knowledge source, we follow the human evaluation setup in \S\ref{subsec:verbalization-eval} to obtain quality annotations of 2000 examples, with 1 annotation per example. This follows a similar setup to \S\ref{sec:purification}--indeed, we are replicating the earlier critic experiments but at a smaller scale (2000 annotations vs. 10000) to allow for more knowledge sources. For each knowledge source, we randomly split into sizes of 1400/300/300 for train, dev, and test sets. We follow \S\ref{sec:purification} to train a critic model for each knowledge source.

\begin{figure}
    \centering
    \includegraphics[width=0.8\linewidth]{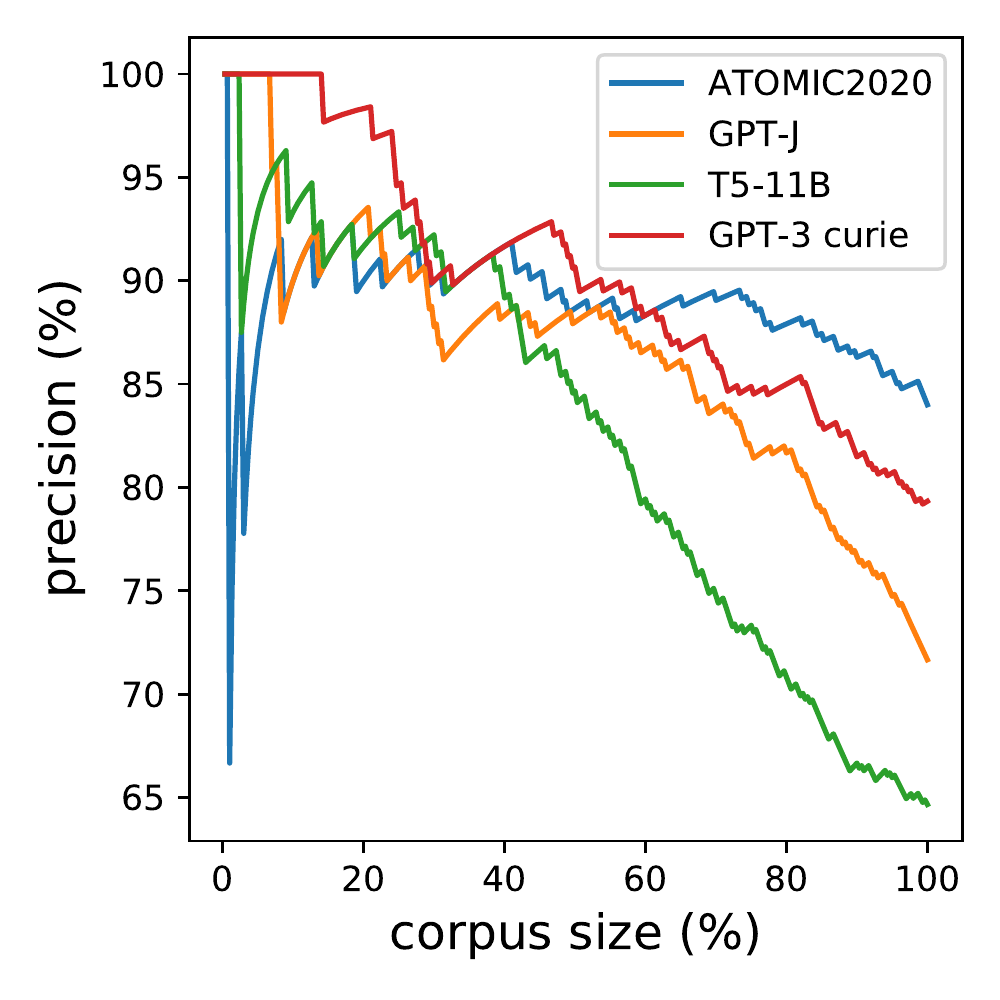}
    \caption{Precision resulting from the critic step from \S\ref{sec:purification}, with various thresholds. We include corpora generated by GPT-3 (\autotomic), GPT-J, T5-11B, and humans (\atomic). Without filtering (corpus size = 1.0), different corpora have a variety of precisions. As more examples are filtered by the critic, precision rises significantly demonstrating the strong value of the critic step.}
    \label{fig:alternate_teacher_prec}
\end{figure}

We plot different thresholds (\% of corpus filtered) against the resulting precision (proportion of corpus that is judged to be ``valid'' knowledge) in Figure~\ref{fig:alternate_teacher_prec}, and give numbers at various sizes in Table~\ref{tab:alternate-knowledge-source}. One striking aspect is that a critic model can raise the precision of any of these knowledge sources to approximately 90\% while retaining 30\% of the original corpus size. While this discards a significant portion of the original generated knowledge, it raises the exciting prospect of using more cost-effective models at a large scale to generate strong commonsense corpora like \autotomic. GPT-J and T5-11B can both be run locally by researchers, unlike GPT-3 which uses a pay-per-generation API. Thus, one can imagine producing a large and high-quality corpus like \autotomic at a lower cost by instead generating a larger volume of knowledge from such an accessible model, and simply filtering to a greater extent.

Another interesting aspect is how the various knowledge sources diverge. Under little to no critical filtering (i.e. corpus size = 1.0), the precision of various knowledge sources is widely spread. Before applying a critic, quality of knowledge source is very important. Indeed, precision is ordered by cost of generation: human \atomic has the highest precision while being the most expensive, followed by GPT-3 (used here) which is pay-per-generation, and finally the two publicly available models. Another point of divergence is for extreme filtering (at approximately 20\% of the original corpus size. All knowledge sources but GPT-3 plateau at approximately 90\% accuracy, while GPT-3 rises towards 100\%. Indeed, this supports our use of GPT-3 in this work, as a high-quality automatic knowledge source. 

\begin{table*}[t]

\begin{tabular}{l|rrrrrrrrrr}
\toprule

& \multicolumn{10}{c}{\textbf{Precision} at \textbf{Corpus Size (\%)}} \\
Knowledge Source & 100 &    90 &    80 &    70 &    60 &    50 &    40 &    30 &    20 &     10 \\
\midrule
\atomic &  84.0 &  86.3 &  87.9 &  89.0 &  88.3 &  88.7 &  91.7 &  90.0 &  90.0 &   90.0 \\
 GPT-J &  71.7 &  76.7 &  81.7 &  83.8 &  86.7 &  88.0 &  88.3 &  87.8 &  93.3 &   90.0 \\
 T5-11B &  64.7 &  66.7 &  70.8 &  74.8 &  79.4 &  84.7 &  89.2 &  92.2 &  91.7 &   93.3 \\
 GPT-3 curie &  79.3 &  81.5 &  85.0 &  86.2 &  88.3 &  90.7 &  91.7 &  90.0 &  98.3 &  100.0 \\
\bottomrule
\end{tabular}

    \caption{Knowledge precision at various corpus sizes (from 100\% to 10\%) based on filtering by the critic model. Precision is calculated by human annotation of valid or invalid knowledge. We consider 4 knowledge sources, as described in Appendix~\ref{app:alternate-teachers}. This corresponds to the data plotted in Figure~\ref{fig:alternate_teacher_prec}.}

    \label{tab:alternate-knowledge-source}

\end{table*}

\begin{figure}
    \centering
    \includegraphics[width=\linewidth]{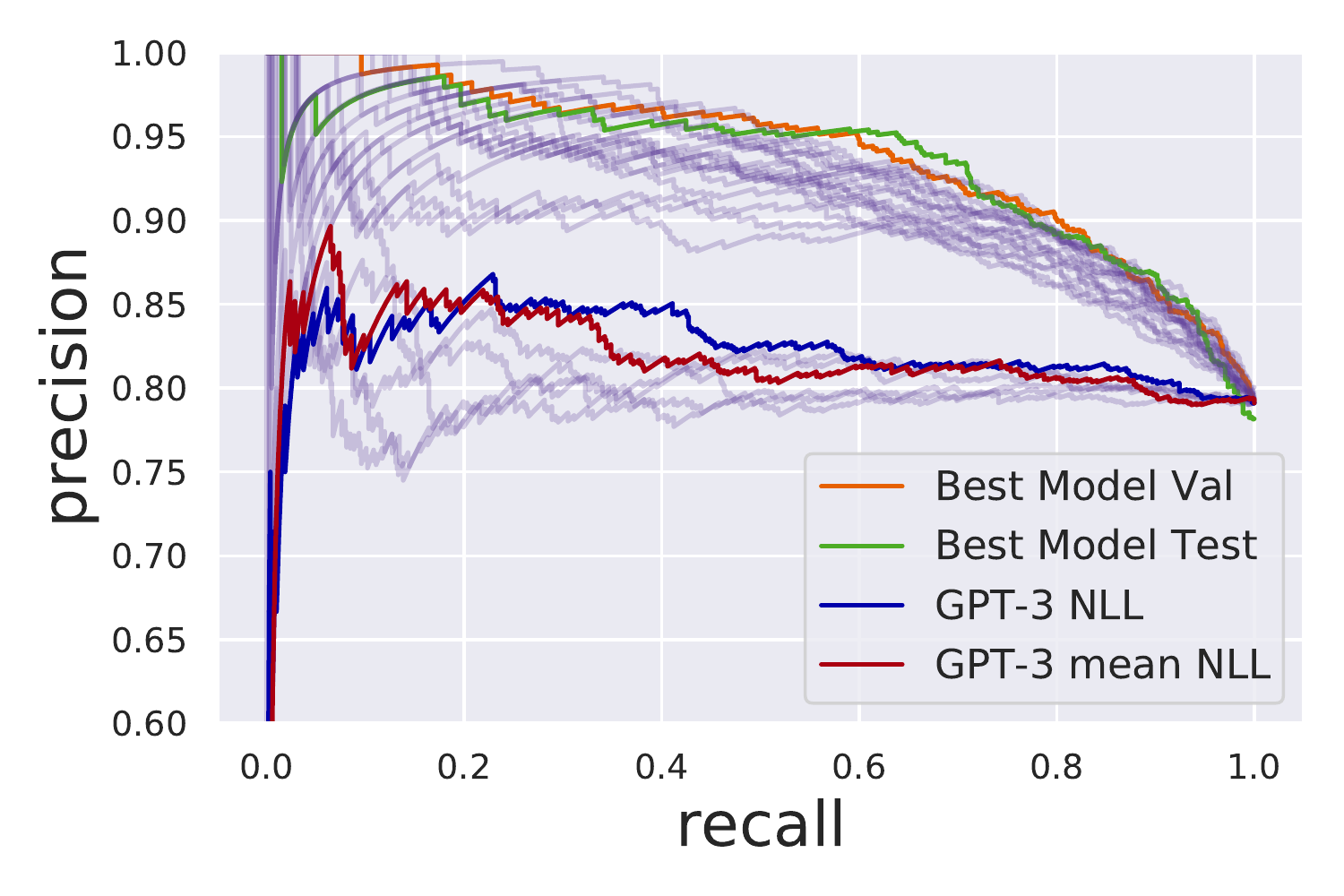}
    \caption{Precision vs. recall of our critic model on the human labelled validation set. The best trained models are labelled, and other hyper-parameter settings are shown as faded lines. We also include generation negative log-likelihood (nll) and token-wise mean nll as cutoff measures--these perform much worse than the supervised model. 
    }
    \label{fig:prec-v-rec}
\end{figure}

\section{Critic Model}

\label{app:critic}

We train binary classifiers (critics) for human acceptability using RoBERTa-Large \cite{liu2019roberta}, fine-tuning all parameters, along with a 2-layer MLP on the \texttt{[CLF]} representation. We conduct a small grid search on the validation set finding batch size 128, dropout .1, and Adam \cite{kingma2014adam} learning rate 5e-6 to be effective. We use early stopping and decay learning rate on validation performance plateauing, to maximize $R@80\%$ on the validation set. We find RoBERTa pretrained on MNLI \cite{N18-1101} effective, outperforming other options. As well, we substitute randomly-sampled names in for person designations ``X''/``Y''. We include as a baseline an unsupervised filtration metric inspired by \cite{davison2019commonsense}: they propose a model estimate of PMI to score mined commonsense triples. In our case, we use Negative Log-Likelihood (NLL) and token-mean-NLL from GPT-3 itself. 

The validation precision/recall of our best performing model, the baselines, and the in-optimal hyperparameter configurations are given in Figure~\ref{fig:prec-v-rec}. Once fixing our model, we applied it to the test set (also in Fig~\ref{fig:prec-v-rec}), verifying that it generalizes to \autotomic entries. Overall, our trained critic model is more effective than the baselines in identifying high and low quality teacher generations at all levels of precision and recall. This result demonstrates that a small amount of human supervision can consistently help to correct GPT-3's mistakes. 

\section{\autotomic Generation Prompts}
\label{app:prompts}

We include example prompts for all generations we do, from Table~\ref{tbl:head_prompt} to \ref{tbl:HinderedBy_prompt}. Note that elements of generation prompts are randomized for each batch. For event generation, the few-shot examples and order are randomly sampled from a seed set of 100 high-quality examples from \atomic in each batch. For inference generation, the natural names used for PersonX and PersonY are randomly sampled from a small predefined set of names.

\begin{table*}[!t]
\setlength{\tabcolsep}{4pt}
\begin{center}
\begin{tabular}{|l|}
\toprule

1. Event: PersonX unwraps PersonY's hands \\ \\ 
2. Event: PersonX overcomes evil with good \\ \\ 
3. Event: PersonX is fed up with the present situation\\ \\ 
4. Event: PersonX breaks PersonX's back \\ \\ 
5. Event: PersonX calls no one \\ \\ 
6. Event: PersonX never gets angry \\ \\ 
7. Event: PersonX does not learn from PersonY \\ \\ 
8. Event: PersonX refuses to touch PersonY's hands \\ \\ 
9. Event: PersonX looks at flowers \\ \\ 
10. Event: PersonX unloads an atomic bomb \\ \\ 
11. Event:
 \\
\bottomrule
\end{tabular}
\end{center}
\caption{Prompt for head generation.}
\label{tbl:head_prompt}
\end{table*}

\begin{table*}[!t]
\setlength{\tabcolsep}{4pt}
\begin{center}
\begin{tabular}{|l|}
\toprule
Next, how are people seen in each situation? Examples:
\\ \\ \\

Situation 1: Devin bullies Jean.
\\ \\
Devin is seen as dominant.
\\ \\
Situation 2: Jamie moves to another city.
\\ \\
Jamie is seen as adventurous.
\\ \\
Situation 3: Sydney changes Ryan's mind.
\\ \\
Sydney is seen as influential.
\\ \\
Situation 4: Lindsay writes a story.
\\ \\
Lindsay is seen as creative.
\\ \\
Situation 5: Rowan covers Pat's expenses.
\\ \\
Rowan is seen as wealthy.
\\ \\
Situation 6: Lee takes time off.
\\ \\
Lee is seen as carefree.
\\ \\
Situation 7: Riley advises Noel.
\\ \\
Riley is seen as informed.
\\ \\
Situation 8: Adrian bursts into tears.
\\ \\
Adrian is seen as depressed.
\\ \\
Situation 9: Hunter deals with problems.
\\ \\
Hunter is seen as responsible.
\\ \\
Situation 10: Sam follows Charlie.
\\ \\
Sam is seen as suspicious.
\\ \\
Situation 11: Alex makes Chris wait.
\\ \\
Alex is seen as \\
\bottomrule
\end{tabular}
\end{center}
\caption{Prompt for generating xAttr.}
\label{tbl:xAttr_prompt}
\end{table*}

\begin{table*}[!t]
\setlength{\tabcolsep}{4pt}
\begin{center}
\begin{tabular}{|l|}
\toprule
Next, what do situations make people do? Examples:

\\ \\ \\
Situation 1: Devin gets a divorce.
\\ \\
As a result, Devin dates someone new.
\\ \\
Situation 2: Jamie lifts weights.
\\ \\
As a result, Jamie has sore muscles.
\\ \\
Situation 3: Sydney takes Ryan to a bar.
\\ \\
As a result, Sydney gets drunk.
\\ \\
Situation 4: Lindsay decides to hire a tutor.
\\ \\
As a result, Lindsay gets better grades.
\\ \\
Situation 5: Rowan buys Pat drinks.
\\ \\
As a result, Rowan is thanked by Pat.
\\ \\
Situation 6: Lee hears bad news.
\\ \\
As a result, Lee begins to cry.
\\ \\
Situation 7: Riley buys a chocolate bar.
\\ \\
As a result, Riley gets change.
\\ \\
Situation 8: Adrian does a lot of work.
\\ \\
As a result, Adrian gets mental fatigue.
\\ \\
Situation 9: Hunter attends a concert.
\\ \\
As a result, Hunter hears a new song.
\\ \\
Situation 10: Sam gets the job done.
\\ \\
As a result, Sam gets more responsibilities.
\\ \\
Situation 11: Alex makes Chris wait.
\\ \\
As a result, Alex \\
\bottomrule
\end{tabular}
\end{center}
\caption{Prompt for generating xEffect.}
\label{tbl:xEffect_prompt}
\end{table*}

\begin{table*}[!t]
\setlength{\tabcolsep}{4pt}
\begin{center}
\begin{tabular}{|l|}
\toprule
For each situation, describe the intent. Examples:
\\ \\ \\

Situation 1: Devin gets the newspaper.
\\ \\
Devin intends to read the newspaper.
\\ \\
Situation 2: Jamie works all night.
\\ \\
Jamie intends to meet a deadline.
\\ \\
Situation 3: Sydney destroys Ryan.
\\ \\
Sydney intends to punish Ryan.
\\ \\
Situation 4: Lindsay clears her mind.
\\ \\
Lindsay intends to be ready for a new task.
\\ \\
Situation 5: Rowan wants to start a business.
\\ \\
Rowan intends to be self sufficient.
\\ \\
Situation 6: Lee ensures Ali's safety.
\\ \\
Lee intends to be helpful.
\\ \\
Situation 7: Riley buys lottery tickets.
\\ \\
Riley intends to become rich.
\\ \\
Situation 8: Alex makes Chris wait.
\\ \\
Alex intends \\
\bottomrule
\end{tabular}
\end{center}
\caption{Prompt for generating xIntent.}
\label{tbl:xIntent_prompt}
\end{table*}

\begin{table*}[!t]
\setlength{\tabcolsep}{4pt}
\begin{center}
\begin{tabular}{|l|}
\toprule
Next, we will discuss what people need for certain situations. Examples:
\\ \\ \\

1. Before Devin makes many new friends, Devin has to spend time with people.
\\ \\
2. Before Jamie gets a date, Jamie has to ask someone out.
\\ \\
3. Before Sydney changes Ryan's mind, Sydney has to think of an argument.
\\ \\
4. Before Lindsay gets a job offer, Lindsay has to apply.
\\ \\
5. Before Rowan takes a quick nap, Rowan has to lie down.
\\ \\
6. Before Lee tries to kiss Ali, Lee has to approach Ali.
\\ \\
7. Before Riley rides Noel's skateboard, Riley has to borrow it.
\\ \\
8. Before Adrian eats the food, Adrian has to prepare a meal.
\\ \\
9. Before Hunter watches Netflix, Hunter has to turn on the TV.
\\ \\
10. Before Sam has a baby shower, Sam has to invite some friends.
\\ \\
11. Before Alex makes Chris wait, Alex has \\
\bottomrule
\end{tabular}
\end{center}
\caption{Prompt for generating xNeed.}
\label{tbl:xNeed_prompt}
\end{table*}

\begin{table*}[!t]
\setlength{\tabcolsep}{4pt}
\begin{center}
\begin{tabular}{|l|}
\toprule
Next, how do people feel in each situation? Examples:
\\ \\ \\

Situation 1: Devin lives with Jean's family.
\\ \\
Devin feels loved.
\\ \\
Situation 2: Jamie expects to win.
\\ \\
Jamie feels excited.
\\ \\
Situation 3: Sydney comes home late.
\\ \\
Sydney feels tired.
\\ \\
Situation 4: Lindsay sees dolphins.
\\ \\
Lindsay feels joyful.
\\ \\
Situation 5: Rowan causes Pat anxiety.
\\ \\
Rowan feels guilty.
\\ \\
Situation 6: Lee goes broke.
\\ \\
Lee feels embarrassed.
\\ \\
Situation 7: Riley has a drink.
\\ \\
Riley feels refreshed.
\\ \\
Situation 8: Adrian has a heart condition.
\\ \\
Adrian feels scared about their health.
\\ \\
Situation 9: Hunter shaves Avery's hair.
\\ \\
Hunter feels helpful.
\\ \\
Situation 10: Sam loses all of Charlie's money.
\\ \\
Sam feels horrible.
\\ \\
Situation 11: Alex makes Chris wait.
\\ \\
Alex feels \\
\bottomrule
\end{tabular}
\end{center}
\caption{Prompt for generating xReact.}
\label{tbl:xReact_prompt}
\end{table*}

\begin{table*}[!t]
\setlength{\tabcolsep}{4pt}
\begin{center}
\begin{tabular}{|l|}
\toprule
Next, what do people want in each situation? Examples:
\\ \\ \\

Situation 1: Devin mows the lawn.
\\ \\
Devin wants to take a shower.
\\ \\
Situation 2: Jamie is going to a party.
\\ \\
Jamie wants to take an Uber home.
\\ \\
Situation 3: Sydney bleeds a lot.
\\ \\
Sydney wants to go to the ER.
\\ \\
Situation 4: Lindsay works as a cashier.
\\ \\
Lindsay wants to find a better job.
\\ \\
Situation 5: Rowan gets dirty.
\\ \\
Rowan wants to do a load of laundry.
\\ \\
Situation 6: Lee stays up all night studying.
\\ \\
Lee wants to rest.
\\ \\
Situation 7: Riley gets Noel's autograph.
\\ \\
Riley wants to tell some friends.
\\ \\
Situation 8: Adrian sees Taylor's point.
\\ \\
Adrian wants to agree with Taylor.
\\ \\
Situation 9: Hunter leaves Avery's bike.
\\ \\
Hunter wants to keep the bike safe.
\\ \\
Situation 10: Sam wants a tattoo.
\\ \\
Sam wants to find a tattoo design.
\\ \\
Situation 11: Alex makes Chris wait.
\\ \\
Alex wants \\
\bottomrule
\end{tabular}
\end{center}
\caption{Prompt for generating xWant.}
\label{tbl:xWant_prompt}
\end{table*}

\begin{table*}[!t]
\setlength{\tabcolsep}{4pt}
\begin{center}
\begin{tabular}{|l|}
\toprule
Next, what can hinder each situation? Examples:

\\ \\ \\
Situation 1: Devin makes a doctor's appointment,
\\ \\
This is hindered if Devin can't find the phone to call the doctor.
\\ \\

Situation 2: Jamie rubs Wyatt's forehead,
\\ \\
This is hindered if Jamie is afraid to touch Wyatt.
\\ \\

Situation 3: Sydney eats peanut butter,
\\ \\
This is hindered if Sydney is allergic to peanuts.
\\ \\

Situation 4: Lindsay looks perfect,
\\ \\
This is hindered if Lindsay can't find any makeup.
\\ \\

Situation 5: Rowan goes on a run,
\\ \\
This is hindered if Rowan injures her knees.
\\ \\

Situation 6: Lee takes Ali to the emergency room,
\\ \\
This is hindered if Ali has no health insurance to pay for medical care.

\\ \\
Situation 7: Riley spends time with Noel's family,
\\ \\
This is hindered if Noel’s family doesn't like spending time with Riley.

\\ \\
Situation 8: Adrian moves from place to place,
\\ \\
This is hindered if Adrian can't afford to move.
\\ \\

Situation 9: Hunter protests the government,
\\ \\
This is hindered if Hunter is arrested.

\\ \\
Situation 10: Sam has a huge fight,
\\ \\
This is hindered if Sam does not like confrontation.
\\ \\

Situation 11: Alex makes Chris wait,
\\ \\
This is hindered if \\
\bottomrule
\end{tabular}
\end{center}
\caption{Prompt for generating HinderedBy.}
\label{tbl:HinderedBy_prompt}
\end{table*}






\end{document}